\documentclass[letterpaper, 10 pt, conference]{ieeeconf}
\IEEEoverridecommandlockouts                               
\usepackage{amsmath}
\usepackage{amssymb}
\usepackage{booktabs}
\usepackage{graphicx}
\usepackage{multirow}
\usepackage{float}
\usepackage{xcolor}
\usepackage{url}
\makeatletter
\let\NAT@parse\undefined
\makeatother
\usepackage{makecell}
\usepackage{hyperref}
\usepackage[font=small, labelfont=bf]{caption}

\definecolor{BestColor}{HTML}{FFE1DE}
\definecolor{BestTextColor}{HTML}{BD1400}
\definecolor{SecondBestColor}{HTML}{fbfbef}
\definecolor{SecondBestTextColor}{HTML}{9f7040}
\newcommand{\best}[1]{{\setlength{\fboxsep}{0.4pt}\colorbox{BestColor}{\textcolor{BestTextColor}{\textbf{\strut #1}}}}}
\newcommand{\secondbest}[1]{{\setlength{\fboxsep}{0.4pt}\colorbox{SecondBestColor}{\textcolor{SecondBestTextColor}{\strut #1}}}}

\title{Repair Before You Fuse: Frozen-Host Adaptation for Corrupted-but-Present Sensors}

\author{Gia-Huy Thai$^{1}$, Quang-Thinh Ly$^{2}$,
Anh-Minh Phan$^{3}$, and Tuan Dang$^{4}$%
\thanks{$^{1}$Gia-Huy Thai is with the University of Science,
VNU-HCM, Ho Chi Minh City 700000, Vietnam.}%
\thanks{$^{2}$Quang-Thinh Ly is with Michigan State University,
East Lansing, MI 48824, USA.}%
\thanks{$^{3}$Anh-Minh Phan is with the Center for Environmental Intelligence, VinUniversity, Hanoi 100000, Vietnam.}%
\thanks{$^{4}$Tuan Dang is with the University of Arkansas,
Fayetteville, AR 72701, USA.}%
}

\begin{document}
\maketitle

\begin{abstract}
Camera-LiDAR detectors can continue to consume unreliable features even when both sensors remain present, synchronized, and calibrated. We introduce \emph{Boundary Feature Repair} (BFR), a frozen-host adaptation framework that learns task-supervised residual corrections at modality interfaces the detector already consumes. BFR-C repairs each camera feature level read by fusion, whereas BFR-L aligns host-conditioned LiDAR candidates to a selected boundary and routes site-wise innovations relative to the frozen anchor. Their jointly trained composition is BFR-CL. Zero-initialized per-channel scales make every variant an exact detector-level identity before optimization; only the repair modules train, while the encoders, fusion consumer, router, detection head, and host normalization statistics remain fixed. At inference, BFR requires neither clean references, corruption metadata, temporal history, nor online updates. Across the complete 20-corruption, five-severity KITTI-C grid, BFR-C reduces RCE from $14.07$ to $11.92$ on MVX-Net and from $14.29$ to $11.00$ on Focals Conv-F relative to their reproduced frozen baselines. On the latter host, BFR-L raises AP$_{\mathrm{cor}}$ from $73.65$ to $74.48$, while BFR-CL reaches $77.01$ AP$_{\mathrm{cor}}$ and $10.46$ RCE with $86.02$ clean AP. On nuScenes-R, BFR-CL raises the reproduced MoME baseline's mAP robustness ratio from $80.1$ to $81.4$. These results establish boundary repair as a targeted retrofit for corrupted-but-present sensing without retraining the deployed detector.

% \cm{Put the concept Figure on the right side of abstract as soon as possible}

\end{abstract}

\section{Introduction}

A sensor need not disappear to fail. In camera-LiDAR detection, both sensors can remain connected, synchronized, and calibrated while providing degraded observations. Camera images may suffer from blur, noise, or poor illumination, while LiDAR measurements may contain sparse or spurious returns. Their encoders can still produce features with the expected shape, allowing corrupted information to pass through the detector. Modern detectors combine these features at point-, voxel-, or proposal-level interfaces~\cite{xu2018pointfusion,sindagi2019mvxnet,vora2020pointpainting}, in a shared bird's-eye-view (BEV) representation~\cite{liu2023bevfusion}, or through object queries and repeated modality interactions~\cite{bai2022transfusion,yang2022deepinteraction}. The fusion module therefore receives unreliable information even though neither sensor appears to be missing.

\begin{figure}[t]
\centering
\includegraphics[width=0.95\columnwidth]{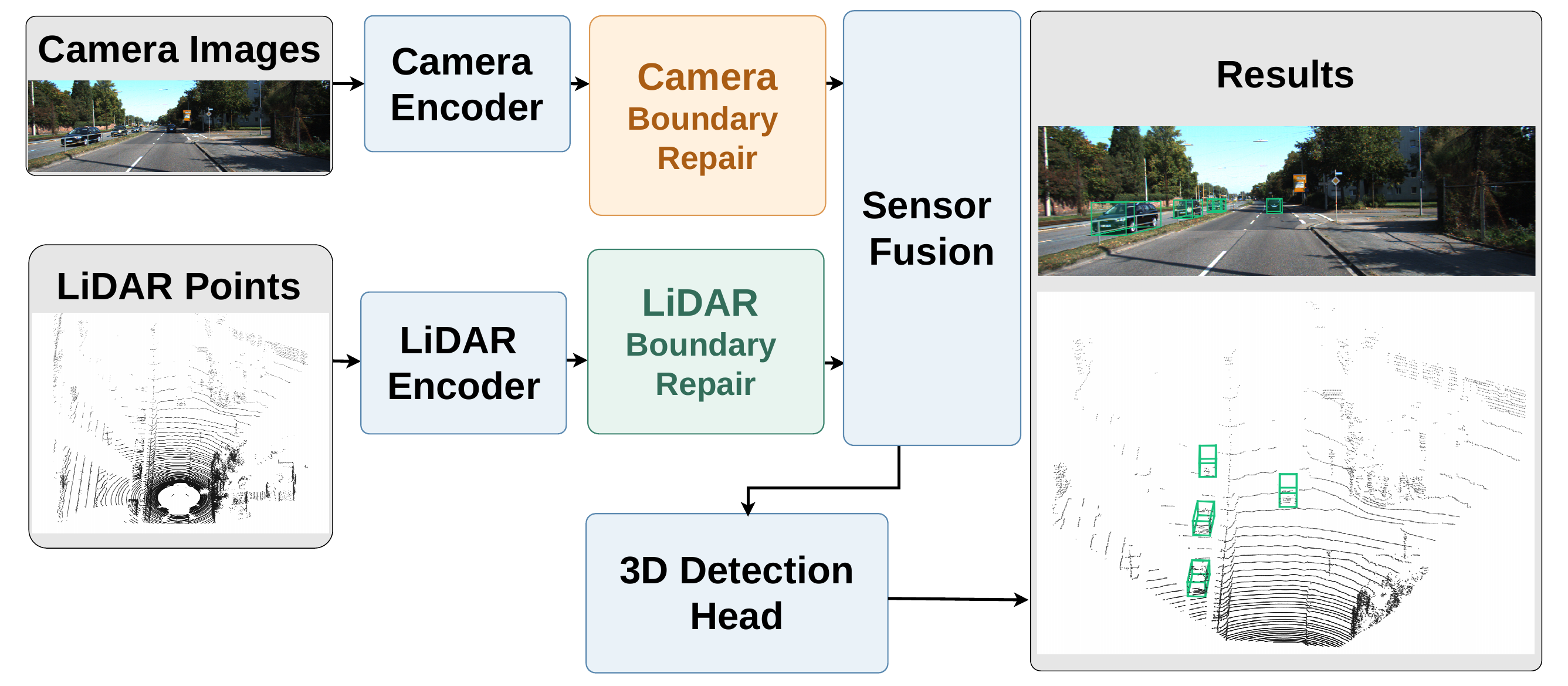}
\caption{Frozen-host boundary repair. BFR-C and BFR-L repair host-consumed camera and LiDAR features while the encoders, fusion consumer, router, and detection head (snowflakes) remain frozen.}
\label{fig:boundary_repair}
\end{figure}

Existing methods address sensor degradation in several ways. Robust fusion approaches modify how a detector denoises, associates, selects, or routes information from different modalities~\cite{bai2022transfusion,ge2023metabev,song2024robofusion,park2025resilient}. Methods for modality loss bypass the missing input or reconstruct its features. ModalPatch, for example, keeps the detector frozen and uses temporal history to predict a temporarily unavailable representation~\cite{li2026modalpatch}. Image restoration, by contrast, aims to recover a clean sensor input. We focus on features that are available in the current frame but corrupted. Our goal is to correct them without retraining the detector or relying on previous observations.

We propose \textbf{B}oundary \textbf{F}eature \textbf{R}epair (BFR), which learns residual corrections at modality-specific interfaces of the frozen detector. The camera branch, BFR-C, applies compact blocks to the final image feature levels used by fusion. At a selected LiDAR boundary, BFR-L constructs host-conditioned candidates, aligns them to the host feature domain, and routes their differences from the frozen anchor independently at each boundary site. This interface-level rule admits representation-derived candidates at a dense BEV boundary and sparse candidates at intermediate LiDAR stages. The combined variant, BFR-CL, applies both repair branches. Figure~\ref{fig:boundary_repair} summarizes their placement in the frozen host.

Zero-initialized channel scales make every BFR variant reproduce the original detector before training. Thereafter, only the repair modules optimize the downstream detection objective; the encoders, fusion consumer, router, detection head, and host normalization statistics remain fixed. BFR thus learns corrections that the frozen consumer can exploit from current evidence rather than reconstructing a generic clean feature. At inference, it requires no corruption labels, clean observations, temporal history, or online updates, but does not assume that completely missing evidence can be recovered.

We evaluate BFR first on KITTI-C~\cite{dong2023benchmarking} and then on nuScenes~\cite{caesar2020nuscenes} under common corruptions and structured sensor failures. On the complete 20-corruption, five-severity KITTI-C grid, both independently trained branches improve corruption AP relative to the same frozen Focals Conv-F host; jointly trained BFR-CL reaches $77.01$ AP$_{\mathrm{cor}}$ with $86.02$ clean AP. BFR-C also transfers to MVX-Net without architectural retuning, reducing RCE by $2.15$ points. On nuScenes-R~\cite{yu2023benchmarking}, BFR-CL improves the reproduced MoME baseline's mAP robustness ratio by $1.3$ points. We report clean accuracy and individual corruption results alongside aggregate scores.

We make three contributions.
\begin{itemize}
    \item We formulate BFR as exact-identity residual repair at modality boundaries of a frozen detector, applied per branch or jointly.

    \item BFR-C places independent residual bottlenecks at every camera level read by pyramid-sampling, sparse-convolution, and query-fusion hosts.

    \item BFR-L aligns host-conditioned candidates at selected dense or sparse LiDAR boundaries and routes site-wise, anchor-relative innovations.
\end{itemize}

\section{Related Work}

\subsection{Fusion Interfaces for Camera-LiDAR Detection}

Fusion architectures differ in how they combine modalities and which features they pass to the downstream detector. Early methods combine separately encoded image and point features at the proposal level~\cite {xu2018pointfusion} or projected points and voxels~\cite{sindagi2019mvxnet}. PointPainting appends image-derived semantic scores to LiDAR points~\cite{vora2020pointpainting}, while EPNet enriches point features with image semantics~\cite{huang2020epnet}. LoGoNet combines local proposal-level and global scene-level fusion~\cite{li2023logonet}, and Focals Conv-F integrates image features through focal sparse convolution~\cite{chen2022focals}. BEVFusion maps both modalities into a shared bird's-eye-view representation~\cite{liu2023bevfusion}. CMT directly processes camera and LiDAR tokens~\cite{yan2023cmt}, UniTR uses a shared multimodal transformer backbone~\cite{wang2023unitr}, and SparseFusion combines sparse candidates in a common 3D space~\cite{xie2023sparsefusion}. TransFusion uses LiDAR-guided object queries to select image evidence~\cite{bai2022transfusion}, while DeepInteraction preserves separate modality representations through repeated cross-modal exchanges~\cite{yang2022deepinteraction}. Although their interfaces differ, these detectors train the fusion components jointly. BFR instead adapts the modality features read by an already-trained fusion consumer while leaving its downstream operations fixed.

\subsection{Robust Fusion under Sensor Degradation}

Robustness benchmarks reveal failures that clean evaluations may overlook. KITTI and nuScenes provide established driving benchmarks~\cite{geiger2012ready,caesar2020nuscenes}. Common-corruption benchmarks extend them with repeatable weather, sensor, motion, object, and alignment perturbations~\cite{dong2023benchmarking}. nuScenes-R focuses on structured camera and LiDAR failures~\cite{yu2023benchmarking}. MultiCorrupt evaluates ten degradation modes and shows that robustness depends strongly on the fusion strategy~\cite{beemelmanns2024multicorrupt}. These benchmarks motivate our evaluation under both sensor corruption and severe sensor failures. Because their corruption recipes, severity levels, checkpoints, and evaluators differ, we report published results as context and base method claims on controlled comparisons with reproduced host baselines.

Existing methods address degradation by changing the detector, its training, or the information available during inference. MetaBEV uses evolving BEV queries to aggregate available modalities~\cite{ge2023metabev}, whereas UniBEV aligns modality-specific BEV maps for operation with different input combinations~\cite{wang2024unibev}. RobBEV combines deformable cross-modal attention with temporal aggregation~\cite{wang2024robbev}. RoboFusion brings together foundation-model image features, frequency-domain denoising, and adaptive fusion~\cite{song2024robofusion}. MoME routes object queries among camera, LiDAR, and joint experts trained to handle sensor failures~\cite{park2025resilient}. RoLiC uses bidirectional feature completion and task-aware distillation for failures affecting one or both modalities~\cite{wang2026rolic}. These approaches train detector components to handle degraded or missing inputs. ModalPatch keeps the detector frozen and predicts temporarily missing features from temporal history, followed by uncertainty-guided refinement~\cite{li2026modalpatch}. BFR also keeps the detector frozen, but corrects degraded features that remain available in the current frame. It requires no temporal history and does not aim to reconstruct a completely missing modality.

\subsection{Frozen-Host Adaptation and Task-Aligned Repair}

Residual learning provides a way to learn corrections around an identity mapping~\cite{he2016deep}. Residual adapters show that compact modules can specialize a largely shared visual network~\cite{rebuffi2017learning}. Feature denoising also modifies internal representations and is trained jointly with an adversarially robust model~\cite{xie2019feature}. BFR trains only the residual repair modules while keeping the complete detector, including its normalization statistics, fixed. Zero-initialized channel scales preserve the original detector's outputs before training. BFR-C applies these corrections to the camera feature pyramid~\cite{lin2017feature}, while BFR-L aligns candidate LiDAR features to a selected host boundary and mixes their innovations relative to its frozen anchor.

Test-time adaptation addresses distribution shift by updating a deployed model using target data. Tent adapts normalization parameters through entropy minimization~\cite{wang2021tent}. MOS maintains and combines historical LiDAR-detector checkpoints online as corruptions change~\cite{chen2025mos}. BFR is trained offline and requires neither optimization on target batches nor a checkpoint bank at inference. Rather than imposing a clean-feature reconstruction target, BFR optimizes directly through the frozen detector's task loss and evaluates performance via downstream detection. This task-aligned objective and offline deployment distinguish it from both feature reconstruction and online adaptation.

\section{Boundary Feature Repair}

% \begin{figure}[t]
% \centering
% \includegraphics[width=\columnwidth]{figures/ICRA'27-OVERVIEW.pdf}
% \caption{Frozen-host boundary repair. BFR-C and BFR-L modify the last camera and LiDAR representations before the fixed multimodal consumer; the encoders, consumer, and detection head (snowflakes) remain frozen. The repaired features follow the unchanged host path to detection. \cm{Revise text in the figure: text size is similar to text size of the caption and readable at size A4}}
% \label{fig:boundary_repair}
% \end{figure}

\begin{figure*}[t]
\centering
\includegraphics[width=0.90\textwidth]{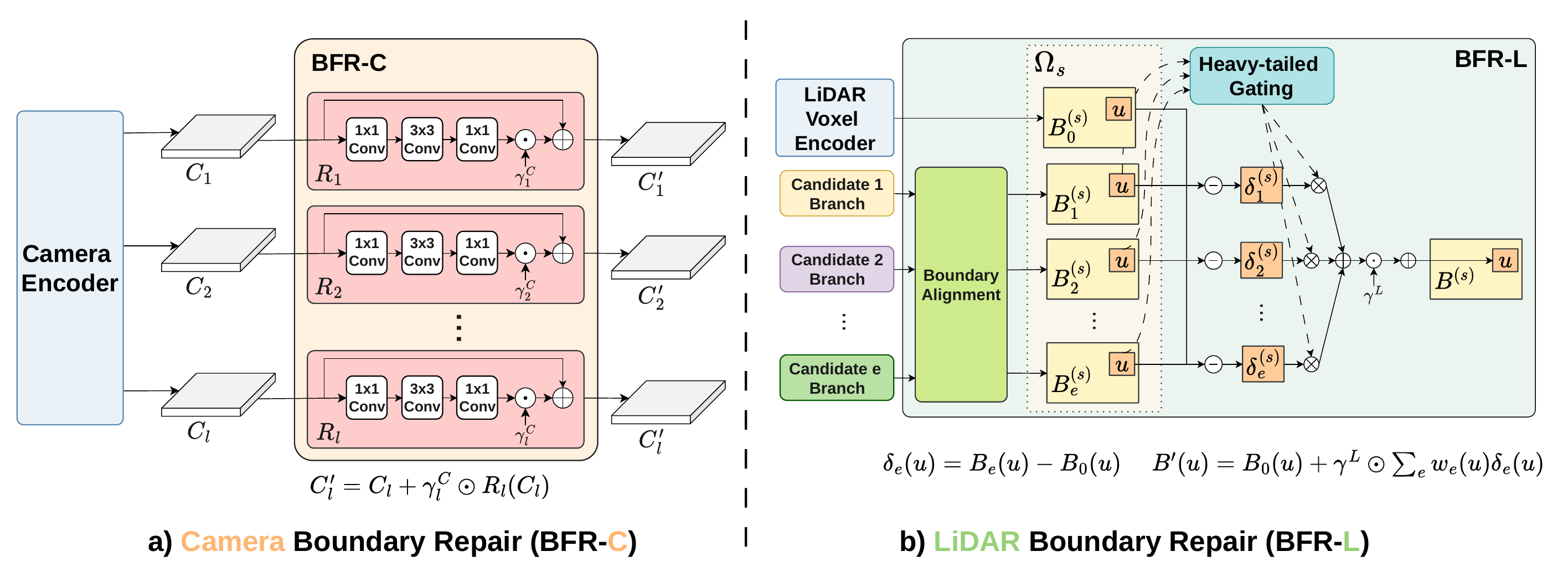}
\caption{Boundary repair mechanisms. (a) BFR-C applies an independent project--spatial--restore residual $R_l$ to every camera feature level consumed by the host, then gates the update with a zero-initialized channel scale $\gamma_l^C$. (b) BFR-L aligns BEV-adapter, point-, and range-derived candidates $B_a$, $B_p$, and $B_r$, routes their anchor-relative innovations with a learned-center heavy-tailed gate, and writes the mixture through $\gamma^L$. The sensor encoders remain frozen. }
\label{fig:bfr-mechanisms}
\end{figure*}

% \cm{Put the overview of the system here: We need a lot of time to modify them since they are the two most important figures }

\subsection{Frozen Consumer, Two Repair Boundaries}
Let $E_c$ and $E_l$ be pretrained camera and LiDAR encoders, $G$ their fusion consumer, and $H$ the detection head. The frozen host maps a camera-LiDAR pair $(x_c,x_l)$ to
\begin{equation}
    \hat{y}_0=H\!\left(G\!\left(B_0,C_{1:L}\right)\right),\quad
    B_0=E_l(x_l),\quad C_{1:L}=E_c(x_c).
    \label{eq:frozen-host}
\end{equation}
We consider corrupted-but-present inputs $(\tilde{x}_c,\tilde{x}_l)$ whose shapes, timing association, and calibration remain valid even though their evidence is degraded. BFR follows representations already consumed by the host: BFR-C maps the final camera pyramid $C_{1:L}$ to $C'_{1:L}$, while BFR-L repairs one or more selected LiDAR boundaries and leaves every downstream host operation unchanged. We write $B'$ for the repaired LiDAR representation produced after these boundaries and their frozen suffix, distinguishing it from the unmodified host output $B_0$. Their composition predicts
\begin{equation}
    \hat{y}_{CL}=H\!\left(G\!\left(B',C'_{1:L}\right)\right).
    \label{eq:bfr-joint}
\end{equation}
The encoders, fusion consumer, expert router, detection head, and all host normalization statistics are fixed. We use corruption type and severity only to construct training and evaluation inputs; we do not supply either to the actuators.

\subsection{Camera Boundary Repair (BFR-C)}
Figure~\ref{fig:bfr-mechanisms}(a) expands the camera actuator across the feature levels read by the host. Each level keeps its identity path, while its repair branch uses the same project--spatial--restore pattern with parameters that are independent across levels.
For camera level $l$ with $c_l$ channels, BFR-C applies
\begin{equation}
    \delta_l^C=\gamma_l^C\odot R_l(C_l),\qquad
    C'_l=C_l+\delta_l^C,
    \label{eq:bfr-camera}
\end{equation}
where $\gamma_l^C\in\mathbb{R}^{c_l}$ is broadcast spatially. The lightweight bottleneck is
\begin{equation}
    R_l(C_l)=W_l^{3}\,\sigma\!\left(\operatorname{GN}\!\left(W_l^{2}\,
    \sigma\!\left(\operatorname{GN}\!\left(W_l^{1}C_l\right)\right)\right)\right),
    \label{eq:bfr-camera-block}
\end{equation}
with $1\!\times\!1$, $3\!\times\!3$, and $1\!\times\!1$ convolutions, ReLU $\sigma$, and GroupNorm in the bottleneck. Parameters are independent across the feature levels a host consumes. MoME and MVX-Net use 256-channel inputs and a 64-channel bottleneck; Focals Conv-F uses a compact 16-to-64-channel instance. This preserves every feature shape and calibration operation downstream.

For MVX-Net~\cite{sindagi2019mvxnet,xu2018pointfusion}, BFR-C repairs all five FPN levels immediately before calibrated point sampling. In Focals Conv-F, it repairs the single 16-channel layer1\_feat2d tensor that the focal sparse-convolution module reads. In the MoME host~\cite{park2025resilient}, the same operator is shared across views and applied at both camera levels before query-level fusion. BFR-C therefore adapts different consumers through the same functional rule: repair the final current-frame camera representation, then leave the original fusion path untouched.

\subsection{LiDAR Boundary Repair (BFR-L)}
BFR-L is defined at a selected host boundary $s$, rather than by one fixed LiDAR representation. Let $B_0^{(s)}:\Omega_s\rightarrow\mathbb{R}^{c_s}$ denote the frozen host anchor on spatial support $\Omega_s$. A set of host-conditioned builders produces candidate features $\widetilde B_e$, indexed by $e\in\mathcal E_s$. The boundary-alignment operator $\mathcal A_{e\rightarrow s}$ maps each candidate to the anchor interface,
\begin{equation}
    B_e^{(s)}=\mathcal A_{e\rightarrow s}(\widetilde B_e),\qquad
    B_e^{(s)}:\Omega_s\rightarrow\mathbb{R}^{c_s}.
    \label{eq:bfr-lidar-align}
\end{equation}
Thus, $s$ selects the repaired host boundary, whereas $u\in\Omega_s$ indexes a site shared by the anchor and every aligned candidate. In Figure~\ref{fig:bfr-mechanisms}(b), the blue encoder denotes the frozen host prefix ending at $s$, which may be an intermediate sparse stage or a dense BEV output. The panel shows one site; the operation applies over the full domain.

At site $u$, each aligned feature, including anchor $e=0$, is projected to $z_e^{(s)}(u)$ and compared with a learned center $\alpha_e^{(s)}$. BFR-L uses the polynomial-tail score
\begin{equation}
\begin{aligned}
    a_e^{(s)}(u)&=\beta_e^{(s)}-\log\!\left(1+\lVert z_e^{(s)}(u)-\alpha_e^{(s)}\rVert_2\right),\\
    w_e^{(s)}(u)&=\operatorname{softmax}_{e\in \mathcal E_s}a_e^{(s)}(u).
\end{aligned}
\label{eq:bfr-lidar-gate}
\end{equation}
where $\beta_0^{(s)}$ initializes a preference for the frozen anchor. The polynomial tail decays more slowly with projected distance than an exponential kernel; we treat this as a design choice rather than attributing the overall repair gain to the kernel alone. BFR-L routes innovations relative to the deployed host instead of averaging replacement features:
\begin{equation}
\begin{aligned}
    \delta_e^{(s)}(u)&=B_e^{(s)}(u)-B_0^{(s)}(u),\\
    B'^{(s)}(u)&=B_0^{(s)}(u)+\gamma^L\odot
    \sum_{e\in\mathcal E_s}w_e^{(s)}(u)\delta_e^{(s)}(u).
\end{aligned}
\label{eq:bfr-lidar}
\end{equation}
The candidate builders are instantiated according to the host interface. In the full MoME route, BFR-L is inserted at the dense BEV boundary before the frozen expert router. A range route projects $(x,y,z,$ intensity, time$)$ to a spherical grid, processes it with a compact encoder-decoder, gathers point features, and scatter-maxes them into BEV; a point route applies a shared MLP and BEV scatter; and an adapter route maps the host anchor. Their spatial and channel alignment is represented by $\mathcal A_{e\rightarrow s}$. The innovation gate also exposes $\rho=[\text{normalized entropy},w_0]$ to the existing query route; $\rho$ is an internal feature, not a calibrated sensor-health label. On Focals Conv-F, two sparse routing instances with 16-channel hidden projections instead attach to the 64-channel \emph{x\_conv3} and \emph{x\_conv4} boundaries, preserve their sparse support and SubM indices, and use host-specific sparse candidates. Each instance follows the same aligned, anchor-relative routing and zero-scale writeback in Equation~\ref{eq:bfr-lidar}; Figure~\ref{fig:bfr-mechanisms}(b) depicts one such boundary.

\paragraph{BFR-CL}
The full plug-in comprises both actuators: BFR-C repairs the camera feature pyramid, and BFR-L repairs the LiDAR representation in one jointly trained run on the same frozen host. Each actuator starts from its zero-scale identity, and no host parameter or normalization statistic is updated. BFR-C and BFR-L in the tables use the same modules trained alone.

\subsection{Exact Identity and Learning Objective}
All camera scales $\gamma_l^C$ and installed LiDAR scales $\gamma^L$ are initialized to zero. Consequently, $C'_l=C_l$ and $B'^{(s)}=B_0^{(s)}$ at every repaired boundary before optimization. Every frozen suffix therefore receives exactly its original input, and Equation~\ref{eq:bfr-joint} gives $\hat{y}_{CL}=\hat{y}_0$ up to numerical determinism. This detector-level identity holds for either branch alone and for their composition. It is an initialization property, not a guarantee of zero clean cost after training, which we measure explicitly.

Both branches receive the unchanged host detection loss. For BFR-C, the clean-camera residual anchor is
\begin{equation}
    \mathcal{L}_C = \mathcal{L}_{\mathrm{det}}+\lambda_C m_C\sum_l\frac{\lVert\delta_l^C\rVert_2^2}{\lvert\delta_l^C\rvert},
\label{eq:bfr-camera-loss}
\end{equation}
where $m_C$ indicates a camera-clean sample. We use $\lambda_C=1$ for the reported MoME BFR-CL model and $0$ for the MVX-Net and Focals Conv-F models. In the full dense route, each aligned BFR-L candidate additionally receives $0.5[\operatorname{SmoothL1}(B_e^{(s)},\operatorname{sg}(B_0^{(s)}))+0.5(1-\cos(B_e^{(s)},\operatorname{sg}(B_0^{(s)})))]$ on occupied sites; this candidate-teacher term is not used by the Focals sparse route. Both routes use load-balancing, LayerScale, and clean-LiDAR residual weights of $0.01$, $0.01$, and $1.0$, respectively, with prior-KL weight $0.05$ inside load balancing. Only the repair modules receive gradients, and we remove all detached targets and auxiliary losses at inference. BFR is therefore optimized for utility to the frozen detection consumer rather than against a generic feature-reconstruction target.

\subsection{Optimization and Inference}
We optimize only the repair parameters offline. Frozen host BatchNorm layers remain in evaluation mode; we confine newly introduced normalization to the repair branches. At inference, we remove corruption sampling, clean indicators, distillation targets, and all auxiliary losses. BFR requires no clean twin, corruption classifier, severity input, temporal cache, target-batch optimization, or change to the detection output space. The added computation is reported in Table~\ref{tab:bfr-efficiency}.

\begin{table*}[t]
\caption{KITTI-C Car 3D AP$_{40}$ at moderate difficulty. BFR variants are compared with their reproduced frozen baselines; $^\dagger$ and $^*$ mark benchmark-reported and RoboFusion-author reproduced results, respectively.}
\label{tab:bfr-kittic}
\centering
\scriptsize
\setlength{\tabcolsep}{2.2pt}
\renewcommand{\arraystretch}{0.98}
\resizebox{\textwidth}{!}{%
\begin{tabular}{cc*{11}{c}}
\toprule
\multicolumn{2}{c}{\textbf{Corruption}}
& \multicolumn{5}{c}{\textbf{LC Fusion}}
& \multicolumn{2}{c}{\textbf{MVX-Net}~\cite{sindagi2019mvxnet}}
& \multicolumn{4}{c}{\textbf{Focals Conv-F}~\cite{chen2022focals}} \\
\cmidrule(lr){3-7}\cmidrule(lr){8-9}\cmidrule(lr){10-13}
&
& EPNet~\cite{huang2020epnet}$^\dagger$ & LoGoNet~\cite{li2023logonet}$^*$
& RoboFusion-L~\cite{song2024robofusion} & RoboFusion-B~\cite{song2024robofusion} & RoboFusion-T~\cite{song2024robofusion}
& (repro.) & \textbf{$+$BFR-C}
& (repro.) & \textbf{$+$BFR-C} & \textbf{$+$BFR-L} & \textbf{$+$BFR-CL} \\
\midrule
\multicolumn{2}{c}{None (AP$_{\mathrm{clean}}$)}
& 82.72 & 85.04 & \best{88.04} & \secondbest{87.87} & 87.60 & 78.59 & 79.33 & 85.92 & 85.84 & 86.17 & 86.02 \\
\midrule
\multirow{4}{*}{\textbf{Weather}}
& Snow & 34.58 & 51.45 & \best{85.29} & \secondbest{84.70} & 84.60 & 27.29 & 37.91 & 34.56 & 54.65 & 38.77 & 54.70 \\
& Rain & 36.27 & 55.80 & \best{86.48} & \secondbest{85.54} & 84.79 & 43.85 & 48.95 & 41.99 & 61.70 & 46.82 & 62.17 \\
& Fog & 44.35 & 67.53 & \best{85.53} & 84.00 & \secondbest{84.17} & 63.03 & 68.65 & 45.08 & 61.25 & 47.19 & 60.26 \\
& S.L. & 69.65 & 75.54 & \best{85.50} & \secondbest{85.15} & 84.75 & 73.80 & 75.46 & 80.04 & 81.79 & 81.59 & 82.39 \\
\midrule
\multirow{9}{*}{\textbf{Sensor}}
& Density & 82.09 & 83.68 & \best{85.71} & 84.34 & 84.11 & 77.08 & 78.48 & 84.96 & 85.15 & \secondbest{85.48} & 85.34 \\
& Cutout & 76.10 & 77.17 & \best{83.17} & \secondbest{81.30} & 81.21 & 72.80 & 73.20 & 80.46 & 80.15 & 80.84 & 80.65 \\
& Crosstalk & 82.10 & 82.00 & 84.12 & 82.45 & 83.07 & 77.46 & 78.61 & 85.83 & 85.70 & \best{86.09} & \secondbest{85.96} \\
& Gaussian (L) & 60.88 & 61.85 & 76.56 & 78.32 & 76.52 & 72.20 & 71.73 & 81.45 & 79.86 & \best{82.52} & \secondbest{82.45} \\
& Uniform (L) & 79.24 & 82.94 & 85.05 & 83.04 & 84.11 & 77.37 & 78.27 & 85.21 & 84.75 & \best{85.40} & \secondbest{85.26} \\
& Impulse (L) & 81.63 & 84.66 & 85.26 & 85.06 & 85.46 & 78.32 & 79.12 & 85.98 & 85.58 & \best{86.13} & \secondbest{86.03} \\
& Gaussian (C) & 80.64 & \secondbest{84.29} & 82.16 & \best{84.63} & 82.17 & 70.48 & 75.26 & 81.37 & 80.82 & 79.85 & 81.66 \\
& Uniform (C) & 81.61 & \secondbest{84.45} & 83.30 & \best{85.20} & 83.30 & 75.60 & 77.20 & 83.00 & 83.02 & 83.06 & 83.50 \\
& Impulse (C) & 81.18 & \secondbest{84.20} & 83.51 & \best{84.55} & 82.91 & 70.65 & 75.14 & 80.59 & 80.78 & 78.60 & 81.57 \\
\midrule
\multirow{2}{*}{\textbf{Motion}}
& Moving Obj. & \best{55.78} & 14.44 & 49.30 & 49.12 & \secondbest{49.90} & 25.08 & 25.61 & 28.20 & 28.52 & 29.23 & 29.16 \\
& Motion Blur & 74.71 & \secondbest{84.52} & 84.17 & \best{84.56} & 84.18 & 74.95 & 77.87 & 81.12 & 83.22 & 82.18 & 83.56 \\
\midrule
\multirow{5}{*}{\textbf{Object}}
& Local Density & 76.73 & 78.63 & 83.21 & 82.53 & 83.22 & 75.68 & 77.04 & 83.39 & 83.58 & \best{83.93} & \secondbest{83.80} \\
& Local Cutout & 69.92 & 64.88 & \best{77.22} & 75.27 & 76.23 & 63.21 & 64.69 & 76.70 & 76.24 & \secondbest{77.19} & 77.13 \\
& Local Gaussian & 75.76 & 55.66 & 79.02 & 78.32 & 78.33 & 75.35 & 76.46 & 82.15 & 80.80 & \secondbest{82.97} & \best{83.18} \\
& Local Uniform & 81.71 & 79.94 & 84.69 & 83.70 & 84.37 & 78.12 & 78.80 & 85.03 & 84.71 & \best{85.66} & \secondbest{85.59} \\
& Local Impulse & 82.21 & 84.29 & 85.26 & 85.08 & 85.06 & 78.35 & 79.02 & 85.81 & 85.62 & \best{86.05} & \secondbest{85.94} \\
\midrule
\multicolumn{2}{c}{Average (AP$_{\mathrm{cor}}$)}
& 70.35 & 71.89 & \best{81.72} & \secondbest{81.31} & 81.12 & 67.53 & 69.87 & 73.65 & 76.39 & 74.48 & 77.01 \\
\multicolumn{2}{c}{RCE (\%) $\downarrow$}
& 14.94 & 15.46 & \best{7.17} & 7.46 & \secondbest{7.38} & 14.07 & 11.92 & 14.29 & 11.00 & 13.57 & 10.46 \\
\bottomrule
\end{tabular}}
\end{table*}

\begin{figure*}[t]
\centering
\includegraphics[width=0.99\textwidth]{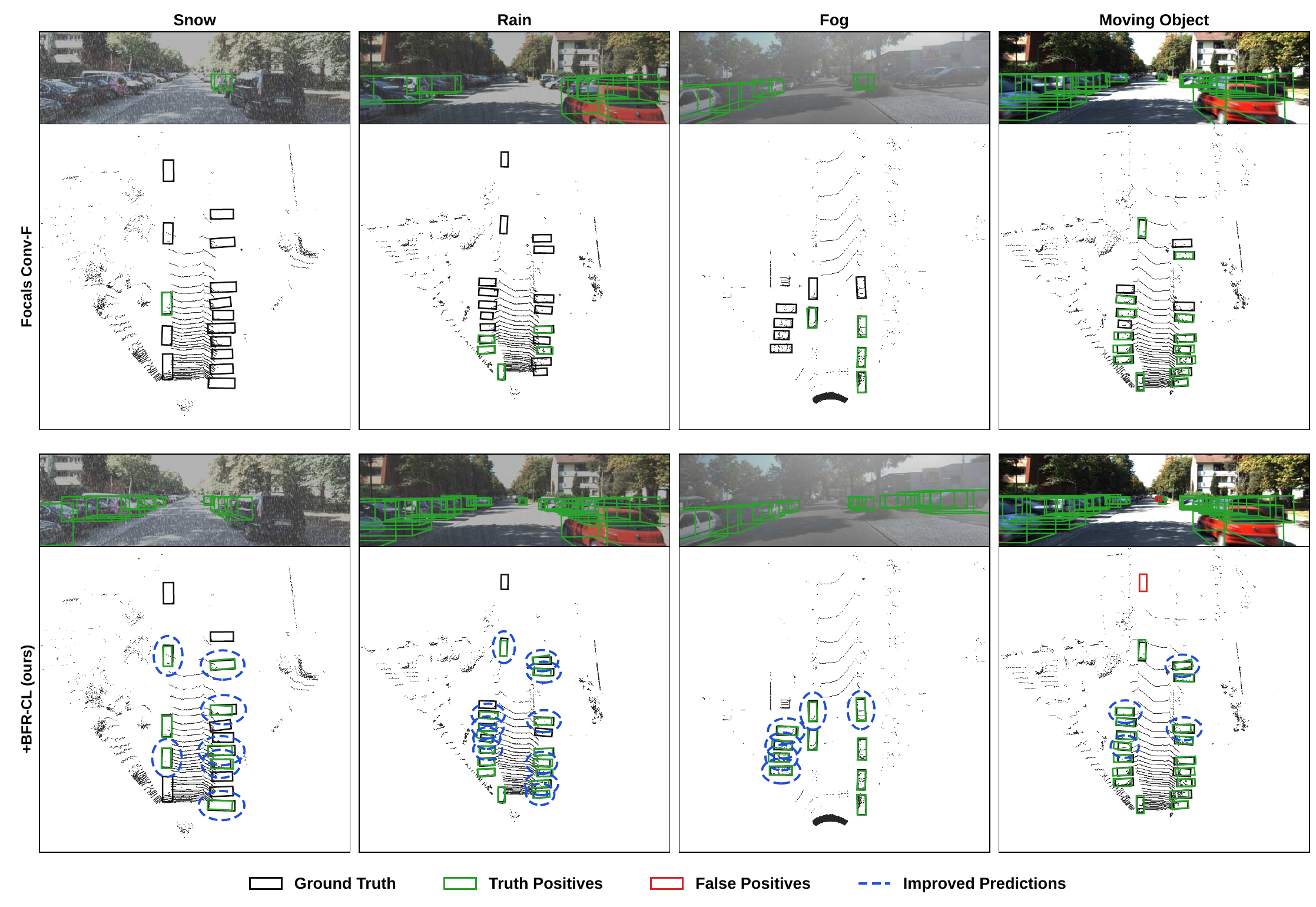}
\caption{KITTI-C severity-2 Car examples for frozen Focals Conv-F (top) and joint BFR-CL (bottom). Black, green, and red denote ground truth, true positives, and false positives; dashed blue circles mark host misses detected by BFR-CL. Both use score $\geq0.5$ and center distance $\leq1.5$~m.}
\label{fig:kittic-qualitative}
\end{figure*}

\section{Experiments}
\subsection{Experimental Setup}
\paragraph{Datasets and metrics}
We evaluate BFR primarily on KITTI-C~\cite{dong2023benchmarking}, reporting Car 3D AP$_{40}$ at moderate difficulty over 20 corruption types and five severities. We define AP$_{\mathrm{cor}}$ as their mean and $\mathrm{RCE}=100(1-\mathrm{AP}_{\mathrm{cor}}/\mathrm{AP}_{\mathrm{clean}})$; RCE is interpreted jointly with clean AP because a lower clean score can reduce it mechanically. As a secondary benchmark, nuScenes~\cite{caesar2020nuscenes} reports mAP and NDS on common corruptions and the nuScenes-R sensor-failure protocol~\cite{yu2023benchmarking}. Severity-2-only aggregates are written AP$_{\mathrm{cor}}$(s2) and RCE(s2). The six nuScenes-R cells follow the official composition and exclude heavier internal variants.

\paragraph{Implementation details}
KITTI-C evaluates BFR on frozen Focals Conv-F~\cite{chen2022focals} and MVX-Net~\cite{sindagi2019mvxnet} hosts. We train BFR-C on Focals Conv-F with AdamW~\cite{loshchilov2019decoupled} for six epochs at learning rate $4.63\times10^{-4}$, corruption probability $0.8$, clean-camera residual weight $\lambda_C=0$, and a 12-family image mix over severities 1--3; two corruptions are composed with probability $0.3$. The locked architecture transfers to MVX-Net without retraining and uses a learning rate of $10^{-3}$. BFR-L on Focals Conv-F trains for six epochs at a learning rate of $2\times10^{-4}$ and a LiDAR corruption probability of $0.5$. Its anchor logit is initialized to $\beta_0=2$, while candidate logits start at zero. BFR-CL jointly trains both modules for six epochs with their inherited optimizer groups and independently sampled camera and LiDAR corruptions; it does not perform joint-specific hyperparameter selection. On nuScenes, the frozen MoME host~\cite{park2025resilient} and normalization statistics remain fixed. BFR-CL trains for six epochs with 8,007 iterations per epoch; camera and LiDAR corruptions are sampled independently with probability $0.3$ at severities 1-3, using the inherited recipe without retuning.

\paragraph{Evaluation protocol}
We compare each BFR result with its corresponding reproduced frozen baseline using the same host checkpoint, evaluator, samples, and corruption realization. Published scores are retained only as context, and only rerun hosts are marked \emph{(repro.)}. Complete five-severity grids use one training seed. We evaluate the nuScenes host and BFR-CL model on the same 6,019 validation samples, the complete severity-2 common-corruption grid, and six nuScenes-R failures.

% \FloatBarrier

\subsection{KITTI-C Evaluation}
\label{sec:complete-benchmarks}

In leaderboard-style comparisons, \colorbox{BestColor}{\textcolor{BestTextColor}{\textbf{Best}}} and \colorbox{SecondBestColor}{\textcolor{SecondBestTextColor}{Second}} mark the best and second-best available values in each displayed row. Protocol differences remain explicit in the captions and must be considered when interpreting these visual rankings.

\paragraph{Complete-grid and cross-host results}
Table~\ref{tab:bfr-kittic} is the principal complete-grid comparison. On Focals Conv-F, BFR-C improves AP$_{\mathrm{cor}}$ by $2.74$ points, from $73.65$ to $76.39$, and reduces RCE by $3.29$ points, from $14.29$ to $11.00$; clean AP changes by only $-0.08$, from $85.92$ to $85.84$. BFR-L independently raises AP$_{\mathrm{cor}}$ by $0.83$ and clean AP by $0.25$, while reducing RCE by $0.72$. Joint BFR-CL gives the largest Focals robustness improvement over the frozen host: AP$_{\mathrm{cor}}$ increases by $3.36$, RCE decreases by $3.83$, and clean AP increases by $0.10$.

Relative to BFR-C alone, joint training adds $0.62$ AP$_{\mathrm{cor}}$ and lowers RCE by another $0.54$ points. The same BFR-C architecture transfers to MVX-Net without retuning: clean AP rises by $0.74$, AP$_{\mathrm{cor}}$ by $2.34$, and RCE falls by $2.15$ points. RoboFusion-L retains the strongest absolute aggregate, with $81.72$ AP$_{\mathrm{cor}}$ and $7.17$ RCE, but it is a separately trained system and therefore provides context rather than a final result. We report clean KITTI performance only through the clean rows paired with KITTI-C; we claim no separate validation table or test-server result.

\paragraph{Per-corruption analysis}
The per-corruption rows reveal complementary, non-uniform responses. BFR-C improves all four weather corruptions on both hosts; on Focals Conv-F, its gains reach $20.09$ AP under snow, $19.71$ under rain, and $16.17$ under fog. Despite these large within-host recoveries, RoboFusion-L retains the highest absolute AP in every weather row. BFR-L has a smaller aggregate effect but improves crosstalk, each of the three LiDAR-noise corruptions, and all five object corruptions; under Gaussian LiDAR noise, it adds $1.07$ AP over the host. Across all displayed methods, BFR-L or BFR-CL ranks first on crosstalk and all three LiDAR-noise rows, leads four of the five object rows, and ranks second on local cutout. Joint BFR-CL combines much of these behavior patterns: compared with BFR-C, it gains $2.59$ under Gaussian LiDAR noise and $2.38$ under local Gaussian corruption, while raising Gaussian-camera performance by $0.84$. It does not dominate every cell---for example, its fog AP is $0.99$ below BFR-C---so the aggregate gain reflects complementary coverage rather than uniform improvement.

\paragraph{Qualitative results}
Figure~\ref{fig:kittic-qualitative} illustrates detection changes behind the aggregate KITTI-C gains. Under snow, rain, and fog, BFR-CL increases the detected ground-truth cars from $1/15$, $5/19$, and $4/10$ to $8/15$, $16/19$, and $10/10$, with no additional false positives in these examples. Under moving-object corruption, it changes $14/18$ to $18/18$ with one additional false positive. These examples illustrate behavior at a fixed threshold rather than estimate recall; the complete-grid evidence is in Table~\ref{tab:bfr-kittic}.

\subsection{nuScenes Evaluation}
We report the jointly trained BFR-CL checkpoint as a system-level intervention and do not assign its gains causally to either branch.

\paragraph{Common corruptions}
Table~\ref{tab:bfr-nuscc} follows the 21-corruption subset of RoboFusion Table~15. The published columns average five severities, whereas we evaluate our reproduced MoME baseline and BFR-CL at severity 2. Rankings across these protocol blocks are therefore contextual. Relative to the reproduced host, BFR-CL improves AP$_{\mathrm{cor}}$(s2) by $1.00$ point, from $62.23$ to $63.23$, and reduces RCE(s2) by $2.07$ points, from $12.58$ to $10.51$; clean mAP decreases by $0.54$. Eleven of 21 corruption cells improve. The gains concentrate under strong sunlight ($+6.56$), crosstalk ($+5.51$), compensation error ($+3.01$), and impulse camera noise ($+2.55$). Several mild density and LiDAR-noise cells decrease by less than $0.4$, so the aggregate improvement is concentrated rather than uniform.

\begin{table}[H]
\caption{nuScenes-C mAP under common corruptions. LC-fusion results average five severities; the reproduced MoME baseline and BFR-CL use severity 2. $^\dagger$ and $^*$ mark published and author-reproduced results.}
\label{tab:bfr-nuscc}
\centering
\scriptsize
\setlength{\tabcolsep}{0.5pt}
\renewcommand{\arraystretch}{0.99}
\resizebox{0.99\linewidth}{!}{%
\begin{tabular}{cc*{5}{c}}
\toprule
\multicolumn{2}{c}{\textbf{Corruption}}
& \multicolumn{3}{c}{\textbf{LC Fusion}}
& \multicolumn{2}{c}{\textbf{MoME}~\cite{park2025resilient}} \\
\cmidrule(lr){3-5}\cmidrule(lr){6-7}
&
& BEVFusion~\cite{liu2023bevfusion}$^\dagger$
& DeepInteraction~\cite{yang2022deepinteraction}$^*$
& RoboFusion-L~\cite{song2024robofusion}
& (repro.)
& \textbf{$+$BFR-CL} \\
\midrule
\multicolumn{2}{c}{\textbf{None} ($\mathrm{mAP}_{\mathrm{clean}}$)}
& 68.45 & 69.90 & 69.91 & \best{71.19} & \secondbest{70.65} \\
\midrule
\multirow{4}{*}{\textbf{Weather}}
& Snow & 62.84 & 62.36 & \best{67.12} & 64.00 & \secondbest{64.72} \\
& Rain & 66.13 & 66.48 & 67.58 & \best{68.63} & \secondbest{68.22} \\
& Fog & 54.10 & 54.79 & 67.01 & \best{68.50} & \secondbest{68.06} \\
& S.L. & 64.42 & 64.93 & \best{67.24} & 60.20 & \secondbest{66.76} \\
\midrule
\multirow{10}{*}{\textbf{Sensor}}
& Density & 67.79 & 68.15 & 69.48 & \best{70.77} & \secondbest{70.39} \\
& Cutout & 66.18 & 66.23 & 69.18 & \best{70.28} & \secondbest{69.90} \\
& Crosstalk & 67.32 & \secondbest{68.12} & \best{68.68} & 10.82 & 16.33 \\
& FOV Lost & 27.17 & 42.66 & 39.48 & \secondbest{51.92} & \best{53.84} \\
& Gaussian (L) & 60.64 & 57.46 & 57.77 & \best{69.97} & \secondbest{69.80} \\
& Uniform (L) & 66.81 & 67.42 & 64.57 & \best{70.67} & \secondbest{70.36} \\
& Impulse (L) & 67.54 & 67.41 & 65.64 & \best{70.54} & \secondbest{70.20} \\
& Gaussian (C) & 64.44 & \secondbest{66.52} & \best{66.73} & 62.12 & 64.42 \\
& Uniform (C) & 65.81 & 65.90 & 65.77 & \secondbest{67.15} & \best{67.32} \\
& Impulse (C) & 64.30 & \best{65.65} & \secondbest{64.82} & 61.59 & 64.14 \\
\midrule
\multirow{2}{*}{\textbf{Motion}}
& Compensation & 27.57 & 39.95 & \secondbest{41.88} & 41.80 & \best{44.81} \\
& Motion Blur & 64.74 & 65.45 & \best{67.21} & 65.70 & \secondbest{65.88} \\
\midrule
\multirow{5}{*}{\textbf{Object}}
& Local Density & \secondbest{67.42} & \best{67.71} & 66.74 & 61.70 & 62.14 \\
& Local Cutout & 63.41 & \secondbest{65.19} & \best{68.82} & 59.79 & 60.46 \\
& Local Gaussian & 64.34 & 64.75 & 65.08 & \best{69.85} & \secondbest{69.68} \\
& Local Uniform & 67.58 & 66.44 & 66.71 & \best{70.59} & \secondbest{70.26} \\
& Local Impulse & 67.91 & 67.86 & 66.53 & \best{70.30} & \secondbest{70.05} \\
\midrule
\multicolumn{2}{c}{\textbf{Average} ($\mathrm{mAP}_{\mathrm{cor}}$)}
& 61.35 & 62.92 & \best{63.90} & 62.23 & \secondbest{63.23} \\
\multicolumn{2}{c}{\textbf{RCE} (\%) $\downarrow$}
& 10.36 & \secondbest{9.97} & \best{8.58} & 12.58 & 10.51 \\
\bottomrule
\end{tabular}}

\end{table}

\begin{table*}[t]
\caption{nuScenes-R robustness under structured sensor failures. $R$ is the mean corrupted-to-clean performance ratio across the displayed failures.}
\label{tab:bfr-nuscr}
\centering
\scriptsize
\setlength{\tabcolsep}{1.8pt}
\renewcommand{\arraystretch}{0.96}
\resizebox{\textwidth}{!}{%
\begin{tabular}{c*{8}{cc}}
\toprule
\multirow{3}{*}{Method}
& \multicolumn{2}{c}{\multirow{2}{*}{Clean}}
& \multicolumn{2}{c}{\multirow{2}{*}{Perf. ratio ($R$)}}
& \multicolumn{8}{c}{LiDAR failures}
& \multicolumn{4}{c}{Camera failures} \\
\cmidrule(lr){6-13}\cmidrule(lr){14-17}
& \multicolumn{2}{c}{} & \multicolumn{2}{c}{}
& \multicolumn{2}{c}{\shortstack{Beam reduction\\\textit{4 beams}}}
& \multicolumn{2}{c}{\shortstack{LiDAR drop\\\textit{all}}}
& \multicolumn{2}{c}{\shortstack{Limited FOV\\\textit{$[-60,60]$}}}
& \multicolumn{2}{c}{\shortstack{Object failure\\\textit{rate $=0.5$}}}
& \multicolumn{2}{c}{\shortstack{View drop\\\textit{6 drops}}}
& \multicolumn{2}{c}{\shortstack{Occlusion\\\textit{w/ obstacle}}} \\
\cmidrule(lr){2-3}\cmidrule(lr){4-5}\cmidrule(lr){6-7}
\cmidrule(lr){8-9}\cmidrule(lr){10-11}\cmidrule(lr){12-13}
\cmidrule(lr){14-15}\cmidrule(lr){16-17}
& mAP & NDS & mAP & NDS & mAP & NDS & mAP & NDS
& mAP & NDS & mAP & NDS & mAP & NDS & mAP & NDS \\
\midrule
UniBEV~\cite{wang2024unibev}$^\dagger$ & 64.2 & 68.5 & 75.8 & 83.7 & 45.1 & 56.1 & 35.0 & 42.2 & 35.9 & 50.1 & 57.1 & 64.0 & 58.2 & 65.3 & 60.5 & 66.4 \\
TransFusion~\cite{bai2022transfusion} & 66.9 & 70.9 & 54.4 & 66.8 & -- & -- & 0.0 & 0.0 & 20.3 & 45.8 & 34.6 & 53.6 & 61.6 & 67.4 & \secondbest{65.5} & \secondbest{70.0} \\
MetaBEV~\cite{ge2023metabev} & 68.0 & 71.5 & 75.4 & 82.5 & -- & 57.7 & 39.0 & 42.6 & -- & \best{57.7} & -- & 67.6 & \best{63.6} & 69.2 & -- & \secondbest{70.0} \\
CMT~\cite{yan2023cmt}$^\dagger$ & 70.3 & 72.9 & 78.4 & 84.4 & \secondbest{54.9} & 62.2 & 38.3 & 44.7 & 43.9 & 54.0 & 66.7 & 70.4 & 61.7 & 68.1 & 65.0 & 69.8 \\
UniTR~\cite{wang2023unitr}$^\dagger$ & 70.5 & \secondbest{73.3} & 54.8 & 68.5 & 47.7 & 59.6 & 0.1 & 1.5 & 22.4 & 48.1 & 38.3 & 55.6 & 60.4 & 67.4 & 62.7 & 68.9 \\
SparseFusion~\cite{xie2023sparsefusion}$^\dagger$ & \secondbest{71.0} & 73.1 & 62.4 & 71.3 & 49.8 & 59.6 & 0.0 & 0.0 & 24.0 & 45.6 & 65.2 & 69.5 & 59.7 & 67.2 & \best{67.2} & \best{71.0} \\
\midrule
MoME~\cite{park2025resilient} (repro.) & \best{71.19} & \best{73.62} & \secondbest{80.1} & \secondbest{85.9} & 54.80 & \secondbest{62.70} & \best{42.40} & \best{47.90} & \secondbest{48.20} & 56.90 & \best{69.50} & \best{72.70} & 63.20 & \best{69.40} & 64.10 & 69.70 \\
\textbf{MoME~\cite{park2025resilient} $+$BFR-CL} & 70.65 & 73.27 & \best{81.4} & \best{86.3} & \best{55.38} & \best{63.04} & \secondbest{42.08} & \secondbest{47.06} & \best{50.50} & \secondbest{57.64} & \secondbest{69.16} & \secondbest{72.45} & \secondbest{63.23} & \secondbest{69.36} & 64.59 & 69.99 \\
\bottomrule
\end{tabular}}

\end{table*}

\paragraph{Structured sensor failures}
Table~\ref{tab:bfr-nuscr} focuses the nuScenes-R comparison on fusion methods. We retain published robustness ratios from their sources; we recompute those for the reproduced MoME baseline and BFR-CL on the six displayed failures and each row's clean score. BFR-CL raises the mAP robustness ratio by $1.3$ points, from $80.1$ to $81.4$, and the NDS ratio by $0.4$, from $85.9$ to $86.3$. Averaged across the six failures, mAP increases by $0.46$, from $57.03$ to $57.49$, whereas NDS changes by only $0.04$, from $63.22$ to $63.26$; clean mAP and NDS decrease by $0.54$ and $0.35$, respectively. Thus, the ratio improvement is accompanied by a modest absolute mAP gain, while the NDS-ratio change partly reflects the lower clean score. Improvements concentrate under limited field of view ($+2.30$ mAP), four-beam LiDAR ($+0.58$), and occlusion ($+0.49$), whereas complete LiDAR drop and object failure decrease by $0.32$ and $0.34$. This contrast delineates the operating regime: boundary repair exploits retained evidence rather than replacing an absent modality.

\subsection{Mechanism Analysis}

\paragraph{BFR-L gate choice}
Table~\ref{tab:bfr-kernel} compares routing rules for the reported sparse Focals BFR-L route. Every entry is the mean AP$_{40}$ change relative to the same reproduced frozen host at severity 2. The eight-corruption subset contains fog, snow, crosstalk, density, cutout, and the three LiDAR-noise corruptions; the 16-corruption set additionally includes rain, strong sunlight, moving-object corruption, and five local object corruptions.

\begin{table}[t]
\caption{Gate-kernel ablation for sparse BFR-L on Focals Conv-F. Entries are severity-2 mean $\Delta$AP$_{40}$ relative to the reproduced frozen host.}
\label{tab:bfr-kernel}
\centering
\scriptsize
\setlength{\tabcolsep}{3.0pt}
\renewcommand{\arraystretch}{0.96}
\resizebox{\linewidth}{!}{%
\begin{tabular}{lrr}
\toprule
Gate kernel & 16 corruptions & 8-corruption subset \\
\midrule
Constant & +1.10 & +0.64 \\
Softmax ($e^{-d^2}$) & +1.12 & +0.68 \\
Laplace ($e^{-d}$) & \secondbest{+1.19} & \secondbest{+0.85} \\
Polynomial tail & \best{+1.27} & \best{+0.90} \\
\bottomrule
\end{tabular}}
\end{table}

All four gates improve over the host on both subsets, showing that Table~\ref{tab:bfr-kernel} selects a routing rule rather than establishing the overall value of BFR-L. The polynomial tail gives the largest point estimates, improving mean AP$_{40}$ by $1.27$ over 16 corruptions and by $0.90$ over the nested eight. Its margins over Laplace are only $0.08$ and $0.05$, respectively. On snow, polynomial-tail and Laplace routing gain $4.83$ and $4.78$ AP, compared with $3.50$ for the Gaussian-distance softmax and $2.96$ for a constant mixture. Given the small margins, we treat the polynomial tail as the reported point-estimate choice rather than a statistically resolved winner. For this route, $\beta_0=2$ initializes the anchor logit two units above the candidate logits and biases early routing toward the frozen feature; the zero-initialized write-back scale $\gamma^L$, rather than this prior, guarantees exact identity.

\paragraph{BFR-C repair response}
Residual probes offer a consistent diagnostic. At the fine KITTI-C FPN level, residual-to-feature amplitude rises from 0.041 on clean input to 0.163 under Gaussian and 0.190 under impulse noise. Under KITTI impulse, energy concentration in image ground-truth regions rises to 1.89 relative to uniform spatial energy. These observations show corruption-responsive structure, but do not themselves establish causal object repair.

\subsection{Efficiency}

\begin{table}[H]
\caption{Model complexity and inference cost. Parameters are totals, and GFLOPs are incremental BFR overhead; dashes denote unavailable measurements.}
\label{tab:bfr-efficiency}
\centering
\scriptsize
\setlength{\tabcolsep}{1.0pt}
\renewcommand{\arraystretch}{0.96}
\resizebox{\linewidth}{!}{%
\begin{tabular}{lccccc}
\toprule
Model / route & Params (M) & Latency (ms) & FPS & VRAM (MiB) & Plug-in GFLOPs \\
\midrule
\multicolumn{6}{l}{\textit{RoboFusion}} \\
RoboFusion-L~\cite{song2024robofusion} & 97.54 & -- & 3.1 & -- & -- \\
RoboFusion-B~\cite{song2024robofusion} & 81.01 & -- & 3.5 & -- & -- \\
RoboFusion-T~\cite{song2024robofusion} & 13.94 & -- & 6.0 & -- & -- \\
\midrule
\multicolumn{6}{l}{\textit{Boundary Feature Repair}} \\
Focals Conv-F~\cite{chen2022focals} (host) & 47.49 & 43.5 & 23.0 & 1009 & -- \\
\quad \textbf{$+$BFR-C} & 47.53 & 42.7 & 23.4 & 1021 & 2.28 \\
\quad \textbf{$+$BFR-L} & 47.55 & 45.6 & 21.9 & 1046 & 6.04 \\
\quad \textbf{$+$BFR-CL} & 47.59 & 47.0 & 21.3 & 1059 & 8.34 \\
\addlinespace[2pt]
MoME~\cite{park2025resilient} (host) & 87.59 & 254.5 & 3.9 & 3901 & -- \\
\quad \textbf{$+$BFR-C} & 87.73 & 253.9 & 3.9 & 3901 & 0.70 \\
\quad \textbf{$+$BFR-L} & 88.30 & 259.8 & 3.8 & 3904 & 46.16 \\
\quad \textbf{$+$BFR-CL} & 88.44 & 260.3 & 3.8 & 3905 & 46.85 \\
\bottomrule
\end{tabular}}
\end{table}

Table~\ref{tab:bfr-efficiency} reports full forward-test latency, including voxelization and post-processing. Because the RoboFusion and BFR rows follow different measurement protocols, their speeds are contextual rather than directly comparable; we interpret overhead within each frozen model. FR-C adds 0.039M parameters, 2.28 GFLOPs, and 12 MiB. Its measured $0.8$ ms reduction is due to timing variation, not evidence of a speedup. BFR-L adds $2.1$ ms and 6.04 GFLOPs, whereas BFR-CL adds $3.5$ ms and 8.34 GFLOPs. On MoME, BFR-CL adds $5.8$ ms and 46.85 GFLOPs. The lightweight efficiency claim therefore applies only to BFR-C; BFR-L and BFR-C apply only to imputation to repair LiDAR features.

% \FloatBarrier

\section{Conclusion}
Boundary Feature Repair adapts corrupted-but-present camera and LiDAR representations at interfaces already consumed by a frozen multimodal detector. BFR-C repairs the camera feature levels supplied to fusion, while BFR-L aligns host-conditioned candidates to selected LiDAR boundaries and routes anchor-relative innovations; zero-initialized scales preserve the complete host at initialization. Controlled comparisons with reproduced baselines on KITTI-C and nuScenes show that these boundary modules can improve corruption robustness while keeping the encoders, fusion consumer, router, detection head, and host normalization statistics fixed. Clean scores and complete-drop results delimit the claim: BFR repairs exploitable evidence in the current observation rather than reconstructing a missing modality.

% \section*{Acknowledgements}
% We gratefully acknowledge \textcolor{red}{text} for supporting this research, and  \textcolor{red}{text} for providing the GPU computing resources essential to this work.
% We will add this section into the the camera ready version

\bibliographystyle{IEEEtran}
\bibliography{refs}

@INPROCEEDINGS{sindagi2019mvxnet,
author={Sindagi, Vishwanath A. and Zhou, Yin and Tuzel, Oncel},
booktitle={2019 International Conference on Robotics and Automation (ICRA)}, 
title={MVX-Net: Multimodal VoxelNet for 3D Object Detection}, 
year={2019},
volume={},
number={},
pages={7276-7282},
doi={10.1109/ICRA.2019.8794195}}

@INPROCEEDINGS{vora2020pointpainting,
author={Vora, Sourabh and Lang, Alex H. and Helou, Bassam and Beijbom, Oscar},
booktitle={2020 IEEE/CVF Conference on Computer Vision and Pattern Recognition (CVPR)}, 
title={PointPainting: Sequential Fusion for 3D Object Detection}, 
year={2020},
volume={},
number={},
pages={4603-4611},
doi={10.1109/CVPR42600.2020.00466}}

@INPROCEEDINGS{liu2023bevfusion,
author={Liu, Zhijian and Tang, Haotian and Amini, Alexander and Yang, Xinyu and Mao, Huizi and Rus, Daniela L. and Han, Song},
booktitle={2023 IEEE International Conference on Robotics and Automation (ICRA)}, 
title={BEVFusion: Multi-Task Multi-Sensor Fusion with Unified Bird's-Eye View Representation}, 
year={2023},
volume={},
number={},
pages={2774-2781},
doi={10.1109/ICRA48891.2023.10160968}}

@InProceedings{bai2022transfusion,
author    = {Bai, Xuyang and Hu, Zeyu and Zhu, Xinge and Huang, Qingqiu and Chen, Yilun and Fu, Hongbo and Tai, Chiew-Lan},
title     = {TransFusion: Robust LiDAR-Camera Fusion for 3D Object Detection With Transformers},
booktitle = {Proceedings of the IEEE/CVF Conference on Computer Vision and Pattern Recognition (CVPR)},
month     = {June},
year      = {2022},
pages     = {1090-1099}
}

@article{yang2022deepinteraction,
title={Deepinteraction: 3d object detection via modality interaction},
author={Yang, Zeyu and Chen, Jiaqi and Miao, Zhenwei and Li, Wei and Zhu, Xiatian and Zhang, Li},
journal={Advances in Neural Information Processing Systems},
volume={35},
pages={1992--2005},
year={2022}
}

@INPROCEEDINGS{park2025resilient,
author={Park, Konyul and Kim, Yecheol and Kim, Daehun and Choi, Jun Won},
booktitle={2025 IEEE/CVF Conference on Computer Vision and Pattern Recognition (CVPR)}, 
title={Resilient Sensor Fusion under Adverse Sensor Failures via Multi-Modal Expert Fusion}, 
year={2025},
volume={},
number={},
pages={6720-6729},
doi={10.1109/CVPR52734.2025.00630}}

@INPROCEEDINGS{caesar2020nuscenes,
author={Caesar, Holger and Bankiti, Varun and Lang, Alex H. and Vora, Sourabh and Liong, Venice Erin and Xu, Qiang and Krishnan, Anush and Pan, Yu and Baldan, Giancarlo and Beijbom, Oscar},
booktitle={2020 IEEE/CVF Conference on Computer Vision and Pattern Recognition (CVPR)}, 
title={nuScenes: A Multimodal Dataset for Autonomous Driving}, 
year={2020},
volume={},
number={},
pages={11618-11628},
doi={10.1109/CVPR42600.2020.01164}}

@INPROCEEDINGS{geiger2012ready,
author={Geiger, Andreas and Lenz, Philip and Urtasun, Raquel},
booktitle={2012 IEEE Conference on Computer Vision and Pattern Recognition}, 
title={Are we ready for autonomous driving? The KITTI vision benchmark suite}, 
year={2012},
volume={},
number={},
pages={3354-3361},
doi={10.1109/CVPR.2012.6248074}}

@inproceedings{dong2023benchmarking,
title={Benchmarking robustness of 3D object detection to common corruptions in autonomous driving},
author={Dong, Yinpeng and Kang, Caixin and Zhang, Jinlai and Zhu, Zijian and Wang, Yikai and Yang, Xiao and Su, Hang and Wei, Xingxing and Zhu, Jun},
booktitle={2023 IEEE/CVF Conference on Computer Vision and Pattern Recognition (CVPR)},
pages={1022--1032},
year={2023},
organization={IEEE}
}

@inproceedings{yu2023benchmarking,
title={Benchmarking the robustness of lidar-camera fusion for 3d object detection},
author={Yu, Kaicheng and Tao, Tang and Xie, Hongwei and Lin, Zhiwei and Liang, Tingting and Wang, Bing and Chen, Peng and Hao, Dayang and Wang, Yongtao and Liang, Xiaodan},
booktitle={2023 IEEE/CVF Conference on Computer Vision and Pattern Recognition Workshops (CVPRW)},
pages={3188--3198},
year={2023},
organization={IEEE}
}

@article{ge2023metabev,
title={Metabev: Solving sensor failures for bev detection and map segmentation},
author={Ge, Chongjian and Chen, Junsong and Xie, Enze and Wang, Zhongdao and Hong, Lanqing and Lu, Huchuan and Li, Zhenguo and Luo, Ping},
journal={arXiv preprint arXiv:2304.09801},
year={2023}
}

@article{rebuffi2017learning,
title={Learning multiple visual domains with residual adapters},
author={Rebuffi, Sylvestre-Alvise and Bilen, Hakan and Vedaldi, Andrea},
journal={Advances in neural information processing systems},
volume={30},
year={2017}
}

@inproceedings{wang2021tent,
title={Tent: Fully Test-Time Adaptation by Entropy Minimization},
author={Dequan Wang and Evan Shelhamer and Shaoteng Liu and Bruno Olshausen and Trevor Darrell},
booktitle={International Conference on Learning Representations},
year={2021},
url={https://openreview.net/forum?id=uXl3bZLkr3c}
}

@InProceedings{xie2019feature,
author = {Xie, Cihang and Wu, Yuxin and Maaten, Laurens van der and Yuille, Alan L. and He, Kaiming},
title = {Feature Denoising for Improving Adversarial Robustness},
booktitle = {Proceedings of the IEEE/CVF Conference on Computer Vision and Pattern Recognition (CVPR)},
month = {June},
year = {2019}
}

@InProceedings{he2016deep,
author = {He, Kaiming and Zhang, Xiangyu and Ren, Shaoqing and Sun, Jian},
title = {Deep Residual Learning for Image Recognition},
booktitle = {Proceedings of the IEEE Conference on Computer Vision and Pattern Recognition (CVPR)},
month = {June},
year = {2016}
}

@InProceedings{lin2017feature,
author = {Lin, Tsung-Yi and Dollar, Piotr and Girshick, Ross and He, Kaiming and Hariharan, Bharath and Belongie, Serge},
title = {Feature Pyramid Networks for Object Detection},
booktitle = {Proceedings of the IEEE Conference on Computer Vision and Pattern Recognition (CVPR)},
month = {July},
year = {2017}
}

@inproceedings{loshchilov2019decoupled,
title={Decoupled Weight Decay Regularization},
author={Ilya Loshchilov and Frank Hutter},
booktitle={International Conference on Learning Representations},
year={2019},
url={https://openreview.net/forum?id=Bkg6RiCqY7},
}

@InProceedings{xu2018pointfusion,
author = {Xu, Danfei and Anguelov, Dragomir and Jain, Ashesh},
title = {PointFusion: Deep Sensor Fusion for 3D Bounding Box Estimation},
booktitle = {Proceedings of the IEEE Conference on Computer Vision and Pattern Recognition (CVPR)},
month = {June},
year = {2018}
}

@article{li2026modalpatch,
  title={ModalPatch: A Plug-and-Play Module for Robust Multi-Modal 3D Object Detection under Modality Drop},
  author={Li, Shuangzhi and Ma, Lei and Li, Xingyu},
  journal={arXiv preprint arXiv:2603.02481},
  year={2026}
}

@inproceedings{song2024robofusion,
  author = {Song, Ziying and Zhang, Guoxing and Liu, Lin and Yang, Lei and Xu, Shaoqing and Jia, Caiyan and Jia, Feiyang and Wang, Li},
  title = {RoboFusion: towards robust multi-modal 3D object detection via SAM},
  year = {2024},
  isbn = {978-1-956792-04-1},
  doi = {10.24963/ijcai.2024/141},
  booktitle = {Proceedings of the Thirty-Third International Joint Conference on Artificial Intelligence},
  articleno = {141},
  numpages = {9},
  location = {Jeju, Korea},
  series = {IJCAI '24}
}

@INPROCEEDINGS{beemelmanns2024multicorrupt,
author={Beemelmanns, Till and Zhang, Quan and Geller, Christian and Eckstein, Lutz},
booktitle={2024 IEEE Intelligent Vehicles Symposium (IV)}, 
title={MultiCorrupt: A Multi-Modal Robustness Dataset and Benchmark of LiDAR-Camera Fusion for 3D Object Detection}, 
year={2024},
volume={},
number={},
pages={3255-3261},
doi={10.1109/IV55156.2024.10588664}
}

@ARTICLE{wang2024robbev,
author={Wang, Jian and Li, Fan and An, Yi and Zhang, Xuchong and Sun, Hongbin},
journal={IEEE Transactions on Circuits and Systems for Video Technology}, 
title={Toward Robust LiDAR-Camera Fusion in BEV Space via Mutual Deformable Attention and Temporal Aggregation}, 
year={2024},
volume={34},
number={7},
pages={5753-5764},
doi={10.1109/TCSVT.2024.3366664}
}

@inproceedings{chen2025mos,
title={{MOS}: Model Synergy for Test-Time Adaptation on Li{DAR}-Based 3D Object Detection},
author={Zhuoxiao Chen and Junjie Meng and Mahsa Baktashmotlagh and Yonggang Zhang and Zi Huang and Yadan Luo},
booktitle={The Thirteenth International Conference on Learning Representations},
year={2025},
url={https://openreview.net/forum?id=Y6aHdDNQYD}
}

@ARTICLE{wang2026rolic,
author={Wang, Lin and Sun, Shiliang and Zhao, Jing},
journal={IEEE Transactions on Image Processing}, 
title={RoLiC: A Robust LiDAR-Camera Fusion Framework for 3D Object Detection}, 
year={2026},
volume={35},
number={},
pages={6846-6859},
doi={10.1109/TIP.2026.3705187}}

@INPROCEEDINGS{yan2023cmt,
author={Yan, Junjie and Liu, Yingfei and Sun, Jianjian and Jia, Fan and Li, Shuailin and Wang, Tiancai and Zhang, Xiangyu},
booktitle={2023 IEEE/CVF International Conference on Computer Vision (ICCV)}, 
title={Cross Modal Transformer: Towards Fast and Robust 3D Object Detection}, 
year={2023},
volume={},
number={},
pages={18222-18232},
doi={10.1109/ICCV51070.2023.01675}}

@INPROCEEDINGS{wang2023unitr,
author={Wang, Haiyang and Tang, Hao and Shi, Shaoshuai and Li, Aoxue and Li, Zhenguo and Schiele, Bernt and Wang, Liwei},
booktitle={2023 IEEE/CVF International Conference on Computer Vision (ICCV)}, 
title={UniTR: A Unified and Efficient Multi-Modal Transformer for Bird’s-Eye-View Representation}, 
year={2023},
volume={},
number={},
pages={6769-6779},
doi={10.1109/ICCV51070.2023.00625}}

@INPROCEEDINGS{xie2023sparsefusion,
author={Xie, Yichen and Xu, Chenfeng and Rakotosaona, Marie-Julie and Rim, Patrick and Tombari, Federico and Keutzer, Kurt and Tomizuka, Masayoshi and Zhan, Wei},
booktitle={2023 IEEE/CVF International Conference on Computer Vision (ICCV)}, 
title={SparseFusion: Fusing Multi-Modal Sparse Representations for Multi-Sensor 3D Object Detection}, 
year={2023},
volume={},
number={},
pages={17545-17556},
doi={10.1109/ICCV51070.2023.01613}}

@INPROCEEDINGS{wang2024unibev,
author={Wang, Shiming and Caesar, Holger and Nan, Liangliang and Kooij, Julian F. P.},
booktitle={2024 IEEE Intelligent Vehicles Symposium (IV)}, 
title={UniBEV: Multi-modal 3D Object Detection with Uniform BEV Encoders for Robustness against Missing Sensor Modalities}, 
year={2024},
volume={},
number={},
pages={2776-2783},
doi={10.1109/IV55156.2024.10588783}}

@inproceedings{huang2020epnet,
  title={Epnet: Enhancing point features with image semantics for 3d object detection},
  author={Huang, Tengteng and Liu, Zhe and Chen, Xiwu and Bai, Xiang},
  booktitle={European conference on computer vision},
  pages={35--52},
  year={2020},
  organization={Springer}
}

@INPROCEEDINGS{chen2022focals,
author={Chen, Yukang and Li, Yanwei and Zhang, Xiangyu and Sun, Jian and Jia, Jiaya},
booktitle={2022 IEEE/CVF Conference on Computer Vision and Pattern Recognition (CVPR)}, 
title={Focal Sparse Convolutional Networks for 3D Object Detection}, 
year={2022},
volume={},
number={},
pages={5418-5427},
doi={10.1109/CVPR52688.2022.00535}}

@inproceedings{li2023logonet,
title={LoGoNet: Towards Accurate 3D Object Detection with Local-to-Global Cross-Modal Fusion},
author={Xin Li and Tao Ma and Yuenan Hou and Botian Shi and Yuchen Yang and Youquan Liu and Xingjiao Wu and Qin Chen and Yikang Li and Yu Qiao and Liang He},
booktitle = {Proceedings of the IEEE Conference on Computer Vision and Pattern Recognition (CVPR)},
year = {2023}
}

\end{document}